\documentclass[conference]{IEEEtran}
\usepackage{cite}
\usepackage{amsmath,amssymb,amsfonts}
\usepackage{algorithmic}
\usepackage{graphicx}
\usepackage{textcomp}
\usepackage{xcolor}
\usepackage[hyphens]{url}
\usepackage{fancyhdr}
\usepackage{hyperref}
\usepackage{xspace}
\usepackage{array}
\usepackage{multirow}
\usepackage{tikz}
\usepackage[normalem]{ulem}
\usepackage{etoolbox}
\usepackage{caption}

\newcommand{\proposed}[0]{SpecMoE\xspace}
\newcommand{\circlednum}[1]{%
  \tikz[baseline=-0.7ex]{\node[draw, circle, fill=black, text=white, inner sep=0.05mm]{#1};}%
  \ignorespacesafterend
}
\robustify{\circlednum}

\newcommand{\fig}[1]{Figure~\ref{#1}}
\newcommand{\equat}[1]{Equation~\ref{#1}}
\newcommand{\sect}[1]{Section~\ref{#1}}
\newcommand{\tab}[1]{Table~\ref{#1}}

\newcommand{\eqn}[1]{Equation~\ref{#1}}

\newcommand{\ondemand}[0]{MoE-OnDemand\xspace}
\newcommand{\overlap}[0]{MoE-Overlap\xspace}
\newcommand{\caching}[0]{MoE-Caching\xspace}
\newcommand{\randomexpert}[0]{Random\xspace}
\newcommand{\hotglobal}[0]{HotGlobal\xspace}
\newcommand{\hottemporal}[0]{HotTemporal\xspace}

\begin{document}

\title{\proposed: A Fast and Efficient Mixture-of-Experts Inference via Self-Assisted Speculative Decoding}

\author{
\IEEEauthorblockN{
Jehyeon Bang\IEEEauthorrefmark{2}\quad
Eunyeong Cho\IEEEauthorrefmark{2}\quad
Ranggi Hwang\IEEEauthorrefmark{4}\IEEEauthorrefmark{1}\quad
Jinha Chung\IEEEauthorrefmark{2}\quad
Minsoo Rhu\IEEEauthorrefmark{2}
}
\IEEEauthorblockA{
\IEEEauthorrefmark{2}KAIST\\
\texttt{\{jehyeon.bang, eunyeong.cho, jinha.chung, mrhu\}@kaist.ac.kr}
}
\IEEEauthorblockA{
\IEEEauthorrefmark{4}UNIST\\
\texttt{ranggi.hwang@unist.ac.kr}
}
}

\maketitle

\begingroup
\renewcommand{\thefootnote}{\fnsymbol{footnote}}
\footnotetext[1]{Work done while at KAIST.}
\endgroup

\begingroup
\renewcommand\thefootnote{}
\footnotetext{This is an extended version of our work, which is accepted for publication at the 63$^{\mathrm{rd}}$ Design Automation Conference (DAC), 2026.}
\addtocounter{footnote}{-1}
\endgroup

\begin{abstract}
The Mixture-of-Experts (MoE) architecture has emerged as a promising approach to mitigate the rising computational costs of large language models (LLMs) by selectively activating parameters.
However, its high memory requirements and sub-optimal parameter efficiency pose significant challenges for efficient deployment.
Although CPU-offloaded MoE inference systems have been proposed in the literature, they offer limited efficiency, particularly for large batch sizes.
In this work, we propose \proposed, a memory-efficient MoE inference system based on our self-assisted speculative decoding algorithm.
\proposed demonstrates the effectiveness of applying speculative decoding to MoE inference without requiring additional model training or fine-tuning.
Our system improves inference throughput by up to 4.30$\times$, while significantly reducing bandwidth requirements of both memory and interconnect on memory-constrained systems.
\end{abstract}

\section{Introduction}

The advent of large language models (LLMs) has transformed language processing, achieving state-of-the-art performance across various tasks. Models like GPT-3~\cite{gpt3}, Llama~\cite{llama}, and Gemini~\cite{gemini} have made significant advancements by scaling up model sizes and increasing training data. However, as model size grows, computational costs increase quadratically with the number of parameters, making large-scale deployment challenging~\cite{scaling_law, llm_scaling_0, llm_scaling_1}. To mitigate this, the Mixture-of-Experts (MoE) architecture~\cite{moe} introduces sparse activation, where only a subset of parameters (known as \emph{experts}) is selected and utilized for each input token. This approach enables MoE models to expand capacity while maintaining computational efficiency, leading to their adoption in high-performance LLMs such as Llama-4~\cite{llama4}, DeepSeek-R1~\cite{deepseekr1}, GPT-4~\cite{gpt4}, and Gemini Pro~\cite{gemini_1.5}.

However, serving MoE models on real systems remains challenging due to their large memory requirements and low parameter efficiency. For example, the increased memory demands of MoE models require additional GPUs just to store model parameters in GPU memory. Despite this, the computational costs of MoE models remain roughly the same as those of their dense counterparts, leading to significant underutilization of GPU computing resources. A common approach to mitigate this inefficiency is to store only a small fraction of model parameters in GPU memory while offloading the rest to slower-tier memory devices, such as CPU DRAM. In this method, model parameters are divided into two groups based on their activation patterns: (1) expert parameters, which are \emph{conditionally} activated and are thus prime candidates for offloading to CPU memory, and (2) non-expert parameters, which are \emph{always} activated in a dense manner and are therefore kept in GPU memory. Although CPU offloading for MoE models reduces the number of GPUs required for inference and helps address GPU resource underutilization, migrating experts from CPU to GPU over the narrow PCIe channel significantly increases end-to-end inference latency.

Previous work primarily employs two strategies to mitigate excessive expert migration latency: (1) overlapping expert migration latency with expert computation or (2) caching frequently accessed experts in GPU memory. The latency-hiding approach~\cite{pregated_moe, se_moe, aptmoe, fastinfmoe} overlaps expert migration latency with expert computation, which is effective when GPU compute latency and CPU-to-GPU communication latency are comparable. However, this method becomes less effective at larger batch sizes ($>$1) because migration latency increases proportionally with batch size, whereas GPU compute latency does not. As a result, migration latency can easily become exposed, especially in batched model-serving scenarios~\cite{vllm, orca}. The caching-based approach~\cite{meta_moe, edgemoe, se_moe, fastinfmoe} partially addresses this issue by analyzing expert access patterns to identify frequently accessed experts, which are then stored in GPU memory. This reduces the number of experts that must be transferred from CPU to GPU memory during inference, thereby decreasing CPU-to-GPU communication over PCIe. However, expert caching faces two key challenges. First, due to limited GPU memory, most expert parameters must remain in CPU memory, leading to a high cache miss rate. Second, in batched inference—an increasingly critical aspect of modern AI model serving—the number of activated experts that must be transferred over PCIe rises significantly as batch size increases, further diminishing the effectiveness of caching.

In summary, the \emph{limitations of prior work primarily stem from its focus on addressing the memory capacity constraints of MoE while neglecting the excessive expert migration latency between the CPU and GPU}. Serving MoE models based on CPU-offloading approaches is fundamentally constrained by the CPU-to-GPU communication bandwidth over PCIe. Thus, reducing the data transfer size (i.e., the volume of expert parameters migrated from CPU to GPU) is key to designing an efficient and scalable MoE inference system.

To this end, we propose \proposed, a high-performance and memory-efficient MoE inference system that utilizes speculative decoding~\cite{blockwise, specdec} to effectively address the memory capacity and communication bandwidth limitations of CPU offloading-based MoE systems. \proposed introduces innovations in both algorithms and MoE inference system design, as detailed below.

\begin{itemize}
    \item \textbf{(Algorithm)} Speculative decoding~\cite{blockwise, specdec} employs a small draft model to generate a sequence of tokens, which the larger target model then verifies before using them as output. However, applying this technique to MoE models presents challenges because MoE models are significantly larger than dense models. Consequently, creating a separate draft model sufficiently large enough to generate high-quality tokens is both impractical and resource-intensive. Our first key contribution is the development of the \emph{self-assisted speculative decoding algorithm}, which leverages parts of the target MoE model as the draft model. This approach eliminates the need to train a separate draft model from scratch, significantly reducing both development costs and the memory required for storage.
    To the best of our knowledge, our work is the first solution that leverages MoE model characteristics to enable speculative decoding for MoE inference.

    \item \textbf{(System)} The \proposed system offers the key advantage of significantly reducing the data transfer size between the CPU and GPU. In conventional MoE systems that rely on CPU offloading, generating an output token requires fetching sparsely activated expert parameters from CPU memory at every token generation step. This process places a heavy demand on communication bandwidth, especially as batch sizes increase. With our self-assisted speculative decoding, multiple output tokens can be generated on average per expert parameter retrieval. This approach effectively reduces the total volume of data transferred over PCIe and helps alleviate the communication bandwidth bottleneck.
\end{itemize}

We implement \proposed on top of an open-source CPU offloading-based inference system and evaluate its performance using representative MoE models. Our results show that \proposed achieves 4.30$\times$ of performance improvement in throughput over other baselines without introducing any additional memory overhead. Furthermore, \proposed significantly reduces CPU-to-GPU data transfer by up to 76.73\%, helping to alleviate the communication bandwidth bottleneck. Overall, \proposed provides a fast and efficient solution for CPU offloading-based MoE inference.

\section{Background and Motivation} \label{sect:background}

\subsection{Mixture-of-Experts Architecture} \label{sect:moe}

\begin{figure}
    \centering
    \includegraphics[width=0.80\columnwidth]{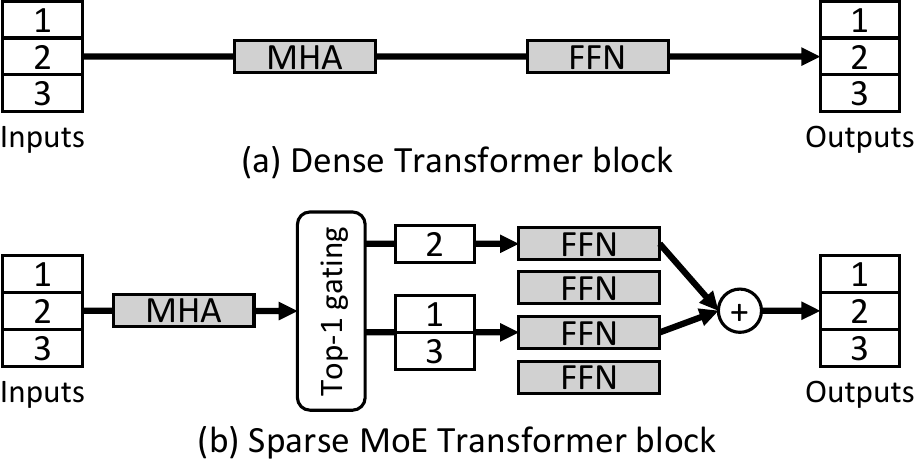}
    \caption{
    (a) Conventional dense Transformer block and
    (b) sparsely activated MoE block.
    The number of activated experts chosen per token ($K$) is set during training, and the model is referred to as a top-$K$ gating model (e.g., top-2 gating in NLLB-MoE~\cite{nllb}).
    }
    \label{fig:moe_arch}
\end{figure}

{\bf Sparsely activated MoE block.}
Transformer-based LLMs generally achieve higher accuracy as model parameters increase~\cite{scaling_law, gpt3, chinchilla, llm_scaling_0}, but their computational cost grows quadratically, making it impractical to design and train these models using conventional methods. To address this scaling challenge, the Mixture-of-Experts (MoE) architecture~\cite{moe, switch, gshard} sparsely activates only a subset of parameters, thereby reducing computational overhead. As shown in \fig{fig:moe_arch}, an MoE Transformer block replaces the feed-forward network (FFN) with a \emph{gate function} and multiple \emph{experts}, each with the same dimension size as the FFN layer. During execution, the gate selects the top-$K$ experts per token, activating only a fraction of the model parameters. This sparse activation strategy significantly increases model capacity compared to conventional dense models while maintaining similar computational requirements.

{\bf Challenges in MoE model inference.}
MoE models typically contain tens of billions of parameters, requiring multiple GPUs for inference~\cite{deepspeed_moe, tutel}. However, their sparse activation leads to significant underutilization of GPU resources. In multi-GPU inference systems, experts are distributed across multiple GPUs using expert parallelism~\cite{switch, gshard}, but only a subset of experts is activated for each inference request. As a result, GPUs holding inactive experts remain idle. This underutilization arises from the low parameter efficiency of MoE models compared to dense models. While MoE models require significantly more memory capacity, their computational demands remain similar to those of dense models.

\begin{figure}[t]
    \centering
    \includegraphics[width=0.99\columnwidth]{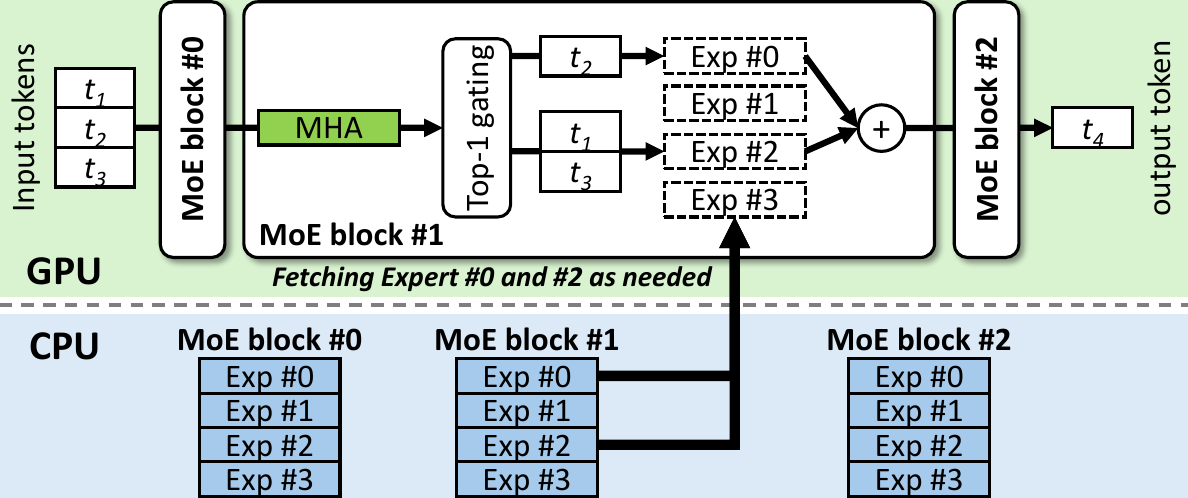}
    \caption{Overview of CPU offloading-based MoE inference system.
    Dense layers are always activated and stored on the GPU, while sparse layers are offloaded to the CPU.
    In this example, three input tokens activate \texttt{Exp~\#0} and \texttt{Exp~\#2}, fetched from CPU memory.
    }
    \label{fig:offloading_system}
\end{figure}

\begin{figure}[t]
    \centering
    \includegraphics[width=0.99\columnwidth]{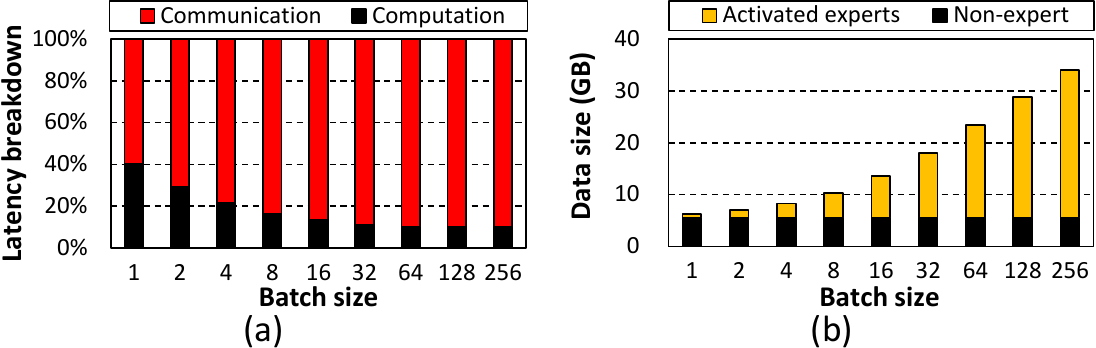}
    \caption{
    (a) Latency breakdown of a single NLLB-MoE's decoding step with different batch sizes using \ondemand on a CPU offloading-based system with a single NVIDIA H100 GPU.
    (b) The size of activated expert parameters vs. non-expert parameters when varying the input batch size in NLLB-MoE.
    The non-MoE parameters (black) are kept in the GPU while the activated experts (yellow) are migrated over the PCIe channel on-demand.
    }
    \label{fig:latency_breakdown}
\end{figure}

One solution is to offload sparsely activated experts to lower memory tiers, such as CPU DRAM or storage, and fetch them into GPU memory on demand (\fig{fig:offloading_system}). This approach (\emph{\ondemand}) improves GPU utilization and reduces the required number of GPUs, thereby lowering inference costs. However, expert migration latency from offloaded devices to the GPU can significantly impact end-to-end latency and severely degrade overall performance. In \fig{fig:latency_breakdown}(a), we present the MoE decoding latency in a CPU offloading-based inference system. As shown, expert migration latency accounts for up to 90.11\% of the total inference latency, highlighting that MoE inference is heavily communication bandwidth-bound. As the number of input tokens increases with larger batch sizes, more experts are activated and fetched from CPU memory. In \fig{fig:latency_breakdown}(b), we compare the data size of non-expert parameters with that of the activated expert parameters. As the size of activated expert parameters grows with larger batches, the CPU-to-GPU communication size increases proportionally. This results in a transfer size exceeding 28 GB per decoding step at a batch size of 256, leading to significant delays due to the narrow PCIe channel. Consequently, limited I/O bandwidth restricts MoE inference scalability and performance.

{\bf Limitation of prior work.} \label{sect:challenge_batch}
Prior work addressing expert migration latency in CPU-offloaded MoE inference can be categorized into two main strategies, each with its unique challenges: (1) compute-communication latency overlap, and (2) expert caching.
\sect{sect:eval} analyzes the applicability and performance of these two prior approaches in greater detail.

\begin{itemize}
\item{\textbf{Compute-communication latency overlap.}}
Prior work has proposed overlapping the communication latency of expert migration with expert computation latency (referred to as \emph{\overlap})~\cite{fastinfmoe, pregated_moe, se_moe}. However, these methods require architectural modifications to the MoE model to enable expert prefetching. This is because expert activation is dynamically determined by the gate function at runtime, just before experts are executed, rendering performance-efficient prefetching of experts infeasible under the conventional MoE architecture.
Such modifications can also reduce model accuracy and necessitate additional training~\cite{pregated_moe}. More critically, \emph{\overlap} is effective only when communication latency is comparable to computation latency. As shown in \fig{fig:latency_breakdown}, communication latency dominates inference latency, accounting for up to 74.8\% at a batch size of 128, severely limiting the effectiveness of latency overlap to small-batch inference. However, small-batch inference is becoming less practical as real-world systems increasingly adopt batched inference for cost-efficient serving (e.g., vLLM~\cite{vllm}, ORCA~\cite{orca}, and Sarathi~\cite{sarathi}).

\item{\textbf{Caching frequently accessed experts.}}
Another line of work has analyzed the expert activation patterns and proposed to \emph{cache} frequently accessed experts in GPU memory to minimize expert migration latency~\cite{meta_moe, edgemoe, fastinfmoe} (referred to as \emph{\caching}). While this approach reduces the number of experts fetched from CPU to GPU, it encounters significant challenges at larger batch sizes. This is particularly true for MoE models with a large number of experts (e.g., NLLB-MoE~\cite{nllb} has 128 experts). Due to the severely limited capacity of GPU memory, only a small fraction of the experts can be cached. Consequently, most experts must be stored in CPU memory, which becomes a critical limitation. In other words, as batch sizes grow, more experts are activated and all must be migrated over the slow PCIe. This limitation makes caching mechanisms ineffective in batched LLM inference.
\end{itemize}

Overall, both \emph{\overlap} and \emph{\caching} only partially mitigate expert migration latency under limited conditions (e.g., single-batch inference) and yield minimal performance gains in practical multi-batch LLM inference scenarios. In summary, they fail to address the critical bottlenecks in CPU offloading-based inference systems: \emph{the frequent and large volume of expert migrations between the CPU and GPU over the narrow interconnect bandwidth}.

\subsection{Speculative Decoding} \label{sect:specdec}

\begin{figure}
    \centering
    \includegraphics[width=0.99\columnwidth]{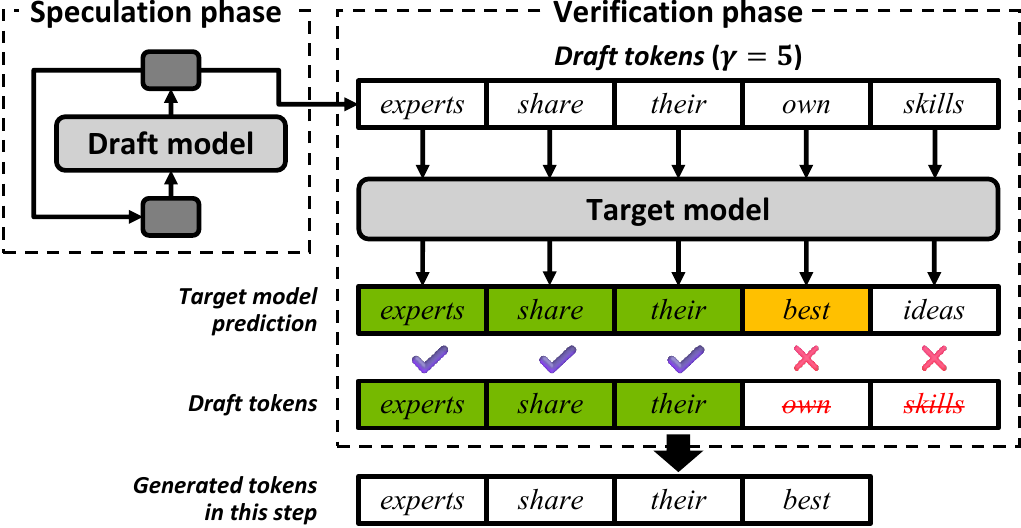}
    \caption{
    Workflow of a speculative decoding step.
    The draft model generates speculated tokens auto-regressively (e.g., 5 tokens in the figure) and feeds them into the target model.
    The earliest rejected token (``own'') and all following tokens are discarded (``own'', ``skills''), while the corrected token (``best'') is appended to the list of accepted tokens.
    In this example, 3 tokens (``experts'', ``share'', ``their'') are accepted and 4 tokens (``experts'', ``share'', ``their'', ``best'') are generated in a single speculative decoding step.
    }
    \label{fig:spec_timeline}
\end{figure}

{\bf Algorithm.}
Speculative decoding is an acceleration technique for LLM inference that reduces end-to-end latency by using a lightweight \emph{draft model} alongside the original \emph{target model}.
Each speculative decoding step consists of two phases: speculation and verification (\fig{fig:spec_timeline}). During the speculation phase, the draft model, which is much smaller and has significantly lower inference latency than the target model, \emph{speculatively} generates several output tokens (the \emph{draft tokens} in \fig{fig:spec_timeline}) in an autoregressive manner. These draft tokens are sent to the target model as inputs for verification during the verification phase, allowing them to potentially serve as the final output tokens.
During the verification phase, the target model determines how many of the speculated tokens can be accepted. It does this by calculating the output logit values of all the input draft tokens \emph{in parallel}, allowing the verification phase to identify which draft tokens will be \emph{accepted} and which should be \emph{rejected}.
During this process, one additional output token is generated (``best'' in \fig{fig:spec_timeline}), which is concatenated to the accepted tokens (``experts'', ``share'', and ``their'') and used as the final output for this decoding step. As a result, a single speculative decoding step can generate \emph{multiple} tokens while loading the entire model's parameters into GPU memory only once.

{\bf Expected speedup.}
\eqn{eq:speedup} shows the expected speedup of speculative decoding. Let $T_{\text{target}}^{<n>}$ represent the latency to process $n$ input tokens in parallel using the target model, where $T_{\text{target}}$ is defined as the latency when $n = 1$ (i.e., $T_{\text{target}} = T_{\text{target}}^{<1>}$). For instance, $T_{\text{target}}^{<5>}$ is the latency of processing the five draft tokens in \fig{fig:spec_timeline}. $c$ denotes the ratio of the draft model's latency $T_{\text{draft}}$ to $T_{\text{target}}$ (s.t. $T_{\text{draft}} = c \cdot T_{\text{target}}$). Since the decoding step is heavily bound by memory bandwidth when loading the model parameters, $T_{\text{target}}^{<n>}$ is approximately equal to $T_{\text{target}}$ for suitable values of $n$. In other words, the latency difference between processing a single draft token and $n$ draft tokens using the target model is negligible.
Let $\tau(\gamma)$ be the average number of tokens generated per speculative decoding step from $\gamma$ draft tokens. Then, to generate $\tau(\gamma)$ tokens, standard decoding executes the target model $\tau(\gamma)$ times, incurring a total latency of $\tau(\gamma) \cdot T_{\text{target}}$. In contrast, speculative decoding takes $\gamma \cdot T_{\text{draft}}$ for speculation plus $T_{\text{target}}^{<\gamma>}$ for verification. Thus, the expected speedup $S$ is:
\begin{align}
\label{eq:speedup}
S = \frac{\tau(\gamma) \cdot T_{\text{target}}}{\gamma \cdot T_{\text{draft}} + T_{\text{target}}^{<\gamma>}} \approx \frac{\tau(\gamma)}{\gamma \cdot c + 1}
\ ( \because T_{\text{target}} \approx T_{\text{target}}^{<\gamma>})
\end{align}

Hence, greater speedup ($S$) can be achieved with (1) a more memory-efficient draft model (i.e., smaller $c$) or (2) a high-quality draft model that generates draft tokens with a higher likelihood of being accepted during the verification phase of speculative decoding (i.e., larger $\tau(\gamma)$). However, these factors often counteract with each other, as the quality of a draft model is closely tied to its model size, making optimal draft model selection challenging.

{\bf Key challenges in applying speculative decoding for MoEs.} \label{sect:challenge_specdec}
Despite its potential, speculative decoding faces two main challenges for MoE inference.
First, unlike the inference of dense LLM models, the latency of an MoE model inference with multiple input tokens is much larger than that of a single input token (i.e., $T_{\text{target}}$ $\not\approx$ $T_{\text{target}}^{<\gamma>}$). This is because having multiple input tokens usually translates into a larger number of experts to be activated, increasing the amount of model parameters fetched from memory. This results in $T_{\text{target}}^{<\gamma>}$ to become multiples of $T_{\text{target}}$, significantly reducing the expected speedup shown in \equat{eq:speedup}.
This problem becomes even worse for CPU offloading-based inference, where expert parameters need to be fetched from lower memory tiers through the slow PCIe channel.
In \fig{fig:latency_breakdown}(b), for example, with 8 input tokens, more than 6 times more experts are activated compared to a single input token on average, leading to roughly 6 times longer expert migration latency.

Furthermore, speculative decoding requires balancing the draft model's acceptance rate with its execution time (i.e., balancing $c$ and $\tau(\gamma)$). Optimizing the draft model's size and quality for these competing demands is non-trivial, making draft model selection a key hurdle in designing MoE-based LLM serving systems with speculative decoding.
This challenge is exacerbated for models with few smaller variants, since the draft model must share the target model's tokenizer.

For these reasons, prior work that tried to apply speculative decoding to MoE inference systems resort to drafter-free speculative decoding methods~\cite{medusa} or reuse parts of the target model to reduce the training cost of the draft model~\cite{eagle, layerskip, speed, draft_verify}.
Nonetheless, these methods still require retraining of some sort, leaving retraining-free solutions for MoE desired.
To the best of our knowledge, no prior work has proposed solutions for effectively designing a draft model for speculative decoding-based MoE inference that avoids the need for a separate MoE model fine-tuning process, facilitating immediate adoption in practical model serving scenarios.

\section{\proposed: An Efficient MoE Inference System using Self-Assisted Speculative Decoding} \label{sect:proposed}

\subsection{High-level Overview} \label{sect:overview}

\begin{figure}
    \centering
    \includegraphics[width=0.96\columnwidth]{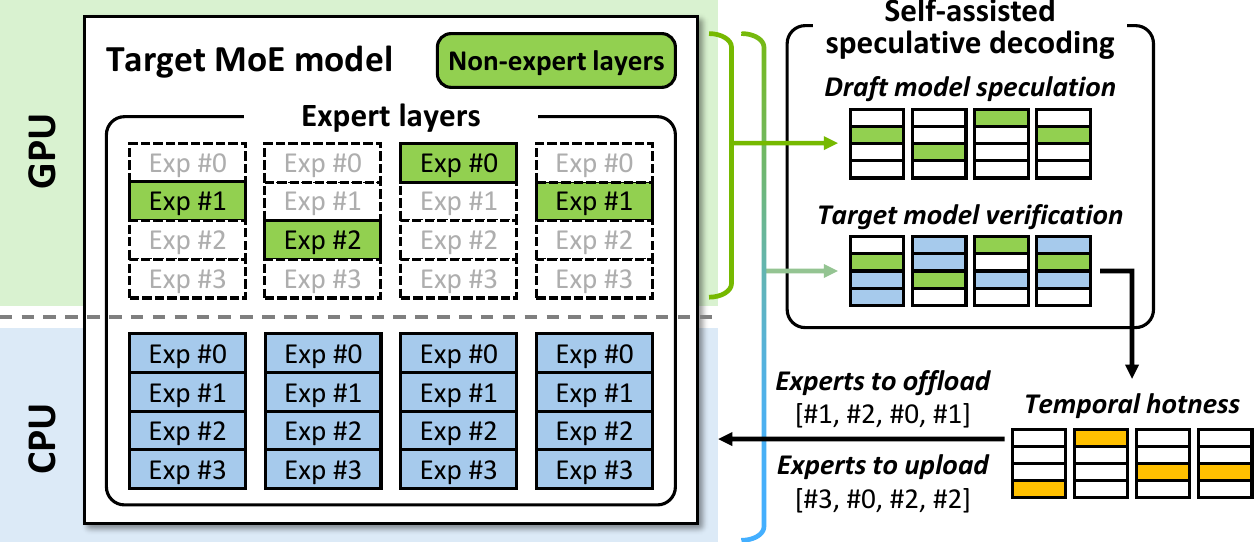}
    \caption{
    High-level overview of \proposed system for a top-1 gating MoE model.
    All non-expert model parameters and a few selected experts are on the GPU (green), while the remaining experts are offloaded to the CPU (blue).
    As a \emph{self-assisted} speculative decoding model, \proposed utilizes parts of the target model (parameters in green) as the draft model.
    With this draft model, whose parameters are entirely on the GPU, speculated draft tokens are generated.
    The target model then verifies the speculated tokens by accessing the required experts, which are fetched from CPU if they are not already on GPU memory (experts in blue).
    To maintain high draft model quality, the experts used to form the draft model are periodically replaced with our proposed expert replacement policy.
    }
    \label{fig:proposed_system}
\end{figure}

\proposed holistically addresses the memory capacity and communication bandwidth challenges of CPU offloading-based MoE inference via speculative decoding. As discussed in \sect{sect:challenge_specdec}, applying speculative decoding to MoE models introduces unique challenges in both the development and optimal deployment of the draft model. We propose a \emph{self-assisted} speculative decoding algorithm that uses parts of the target MoE model as its draft model. This completely eliminates the need for additional training or separate draft model development. \fig{fig:proposed_system} provides an overview of \proposed. We detail the design of our draft model and inference system in the following sections.

\subsection{(Algorithm) Self-Assisted Speculative Decoding} \label{sect:algo}

{\bf \proposed draft model design.}
To create an effective draft model from the original target model, we begin by categorizing the model parameters into two groups: (1) non-expert parameters, which are always used and remain pinned in GPU memory, and (2) expert parameters, which are conditionally activated and offloaded to CPU memory. Since non-expert layers always reside in GPU memory and contribute minimally to latency, we construct our draft model using them, allowing our approach to naturally share its parameters with the target MoE model to ensure high quality.

\begin{figure}
    \centering
    \includegraphics[width=0.97\columnwidth]{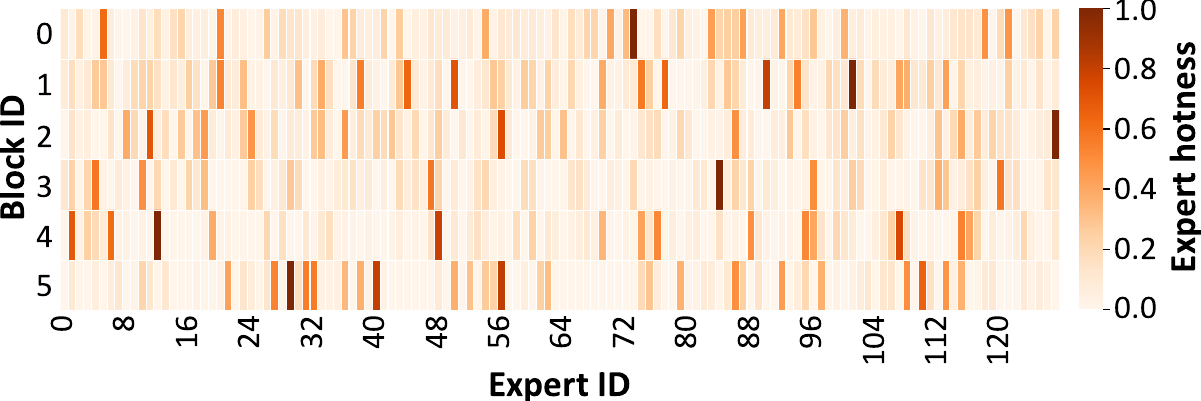}
    \caption{
    Expert hotness of NLLB-MoE in the English-French translation task on the WMT-14 dataset~\cite{wmt}. NLLB-MoE contains 6 MoE blocks, each containing 128 experts. All the MoE blocks within this MoE model (y-axis, showing different MoE block ID) are depicted, showing how different experts within an MoE block (x-axis, showing expert ID within that MoE block) exhibit different hotness.
    }
    \label{fig:hotness}
\end{figure}

Prior work has observed that, during the training phase of MoE models, a subset of experts is routed more frequently than others~\cite{mixtral, meta_moe, nllb}. This results in a skewed distribution of expert activation patterns, referred to as \emph{expert activation hotness}, where certain experts within a single MoE block are more likely to be activated. In \fig{fig:hotness}, we illustrate the expert activation hotness observed in NLLB-MoE, where a small subset of hot experts accounts for a large portion of activations. Based on the observation that hot experts are utilized more frequently and have a greater influence on outputs, we hypothesize that our self-assisted draft model may not require all experts to generate high-quality draft tokens. In other words, by keeping only a few hot experts in GPU memory and routing exclusively to these hot experts, we can create a lightweight version of the target MoE model that performs sufficiently well as a draft model. The intuition behind this approach is that hot experts are routed most of the time, making this new MoE-based draft model an accurate approximation of the original target model.

\begin{figure}
    \centering
    \includegraphics[width=1.0\columnwidth]{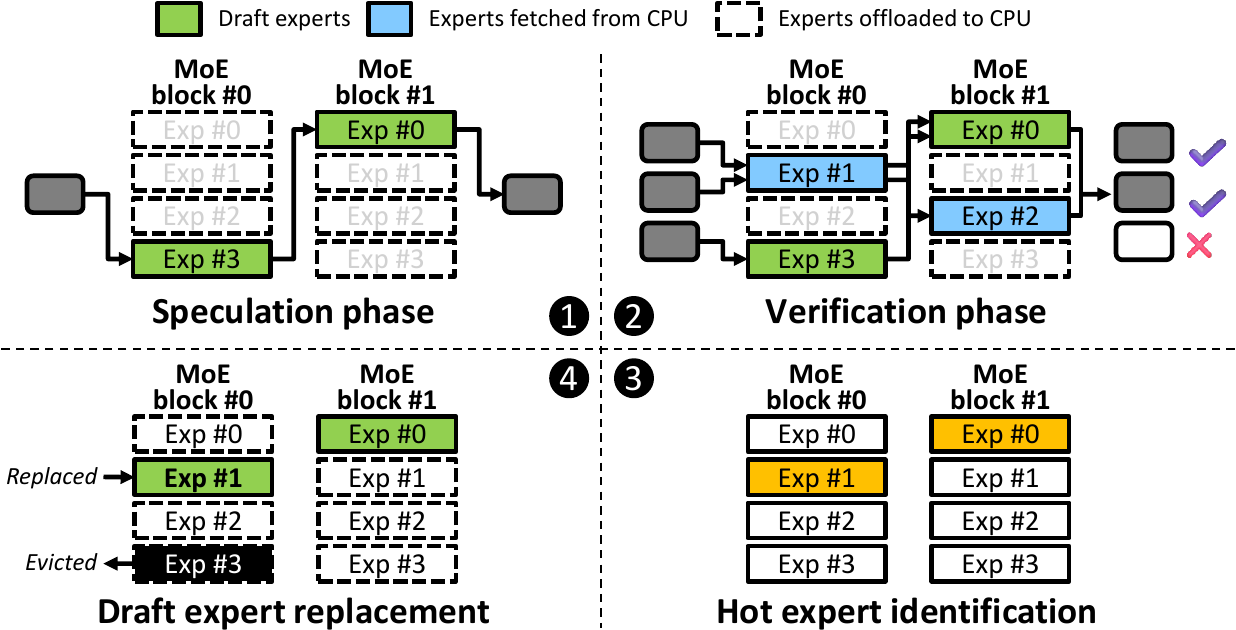}
    \caption{
    The workflow of \proposed's self-assisted speculative decoding algorithm. Example assumes a top-1 gating MoE model.
    \circlednum{1} For each MoE block, only $N$=1 expert is pinned in the GPU as a draft expert and used to generate draft tokens (i.e., draft experts in green).
    \circlednum{2} The draft tokens are then verified by the target MoE model, migrating any required experts not already on the GPU (experts in blue) from the CPU, on demand.
    \circlednum{3} During verification, the temporally hot experts for each MoE block are identified (e.g., \texttt{Exp \#1} of MoE block \#0 is assumed as the most activated expert in this example).
    \circlednum{4} Consequently, the identified $N$ temporally hottest experts are pinned in the GPU as the next draft expert, evicting any draft experts that were not identified as the hottest $N$ experts.
    In this example, the draft experts in MoE block \#0 changes from \texttt{Exp \#3} to \texttt{Exp \#1}, while those in MoE block \#1 remains unchanged.
    }
    \label{fig:specmoe_base}
\end{figure}

Based on these insights, \fig{fig:specmoe_base} provides an illustrative example of \proposed's self-assisted speculative decoding algorithm, highlighting its speculation and verification phases. In this example, the experts statically pinned in GPU memory (\texttt{Exp \#3} for MoE block \#0, and \texttt{Exp \#0} for MoE block \#1) are assumed to be part of the draft model and are referred to as \emph{draft experts}. Because our self-assisted draft model is designed as an MoE architecture, these draft experts serve as a substitute for the MoE blocks in the original target MoE model. Throughout this paper, we refer to \emph{the number of draft experts} pinned in the GPU as $N$, where $N \geq K$ for top-$K$ gating MoE models. Essentially, our self-assisted draft model becomes a smaller MoE model in itself, routing to $K$ experts out of $N$ experts in each MoE block. With this design, no experts are migrated from CPU to GPU memory during the speculation phase, as the activated experts are exclusively those already pinned in GPU memory as part of the draft experts. This approach leverages natural expert activation hotness to maintain high draft model quality and eliminates the cost of training a separate draft model from scratch.

{\bf Expert replacement policy.}
Given the limited GPU memory capacity, the decision regarding which experts to choose for the draft model and pin in GPU memory is critical.
To maintain high quality of the draft tokens generated by the draft model, \proposed's self-assisted speculative decoding algorithm also includes an expert replacement mechanism. \proposed's replacement policy leverages the temporal locality of expert activation patterns~\cite{mixtral, edgemoe} to determine which experts should remain pinned on the GPU to be utilized as the next speculative decoding step's draft experts. We observe that, across different inference runs, there is a temporal shift in expert activation patterns. Specifically, even though some experts are activated more frequently than others, the degree of expert hotness and the skewness of the activation pattern gradually shift during model inference.

To fully leverage these characteristics, \proposed monitors expert activations during the verification phase to identify the temporally hot experts.
At the end of the verification phase, it selects which experts to pin as draft experts for the next decoding step based on their temporal locality.
\fig{fig:specmoe_base} provides an example of how expert replacement policy is employed in \proposed. During the verification phase, \proposed counts which experts were activated the most for the speculated tokens in that round.
For example, in MoE block \#0, \texttt{Exp \#3} was the draft expert during the previous speculation phase, but now \texttt{Exp \#1} emerges as the top-$N$ expert ($N$=1) as a temporally hot expert. Consequently, \proposed replaces the previously pinned \texttt{Exp \#3} with \texttt{Exp \#1}, while in MoE block \#1, \texttt{Exp \#0} remains unchanged for the next speculation phase.
Importantly, this replacement incurs no extra CPU-to-GPU migration since the new draft experts are already in GPU as they have been brought from CPU during the verification phase (i.e., \texttt{Exp \#1} of MoE block \#0 was brought into GPU memory during verification, as indicated by the blue-colored \texttt{Exp \#1} in step \circlednum{2}).
By periodically refreshing the pinned experts to match shifting activation patterns, \proposed dynamically enhances draft quality and maximizes speedup.
As discussed in \sect{sect:low_hotness}, this locality-aware replacement is effective even for MoE models with relatively low expert hotness (e.g., Mixtral-8x7B~\cite{mixtral}).

\begin{figure}
    \centering
    \includegraphics[width=0.99\columnwidth]{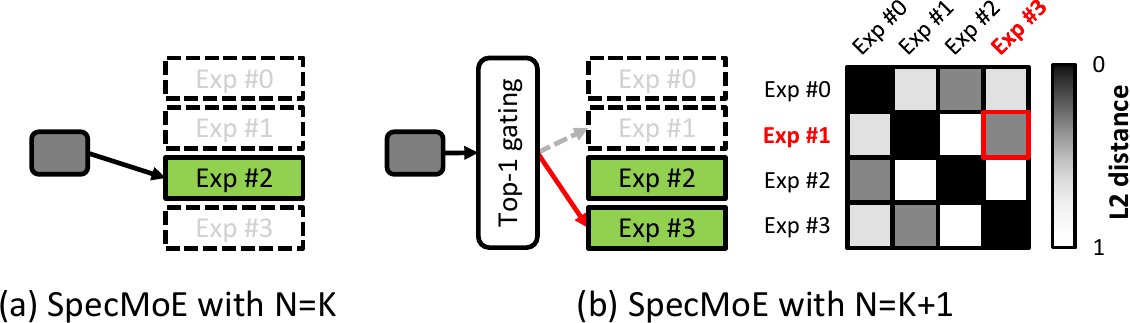}
    \caption{Affinity-based expert selection mechanism in the \proposed speculation phase. (a) Expert selection when the number of draft experts ($N$) equals to the number of routed experts per token ($K=1$). (b) When $N>K$, each token is routed to the most suitable $K$ experts among $N$ draft experts by referring to the affinity table.}
    \label{fig:advanced}
\end{figure}

{\bf Affinity-based expert selection mechanism.}
Since \proposed utilizes the target model's gate function as the draft model's gate function as-is, without fine-tuning, some tokens may be routed to experts that are not currently among the draft experts stored in GPU memory.
To address this, we adopt an \emph{affinity-based expert selection mechanism} illustrated in \fig{fig:advanced}.
Let $N$ be the number of draft experts pinned in GPU memory (with $N \geq K$). When $N=K$, as shown in \fig{fig:advanced}(a), all draft experts are used in the draft model computation. However, if $N > K$, the gating function may select an expert that is not pinned in GPU memory. In that case, \proposed consults an affinity table to find the draft expert that best approximates the target expert. The affinity table contains the precomputed similarity of all possible expert pairs among the experts in the target model using L2 distance. For example, as illustrated in \fig{fig:advanced}(b), if the gating function selects an expert that is not a draft expert (e.g., \texttt{Exp \#1}), \proposed checks the affinity table to identify the draft expert that most closely resembles it (e.g., \texttt{Exp \#3}) and routes the token to that expert instead.
The precomputed affinity table requires only an offline, one-time cost and uses minimal memory (under 200KB for NLLB-MoE with 128 experts per block), all while preserving token quality by closely approximating the original expert's response.
Overall, this mechanism enables \proposed to maintain high token acceptance rates and robust speculative decoding across diverse MoE models, even when hotness patterns are less pronounced.

\begin{figure}
    \centering
    \includegraphics[width=0.96\columnwidth]{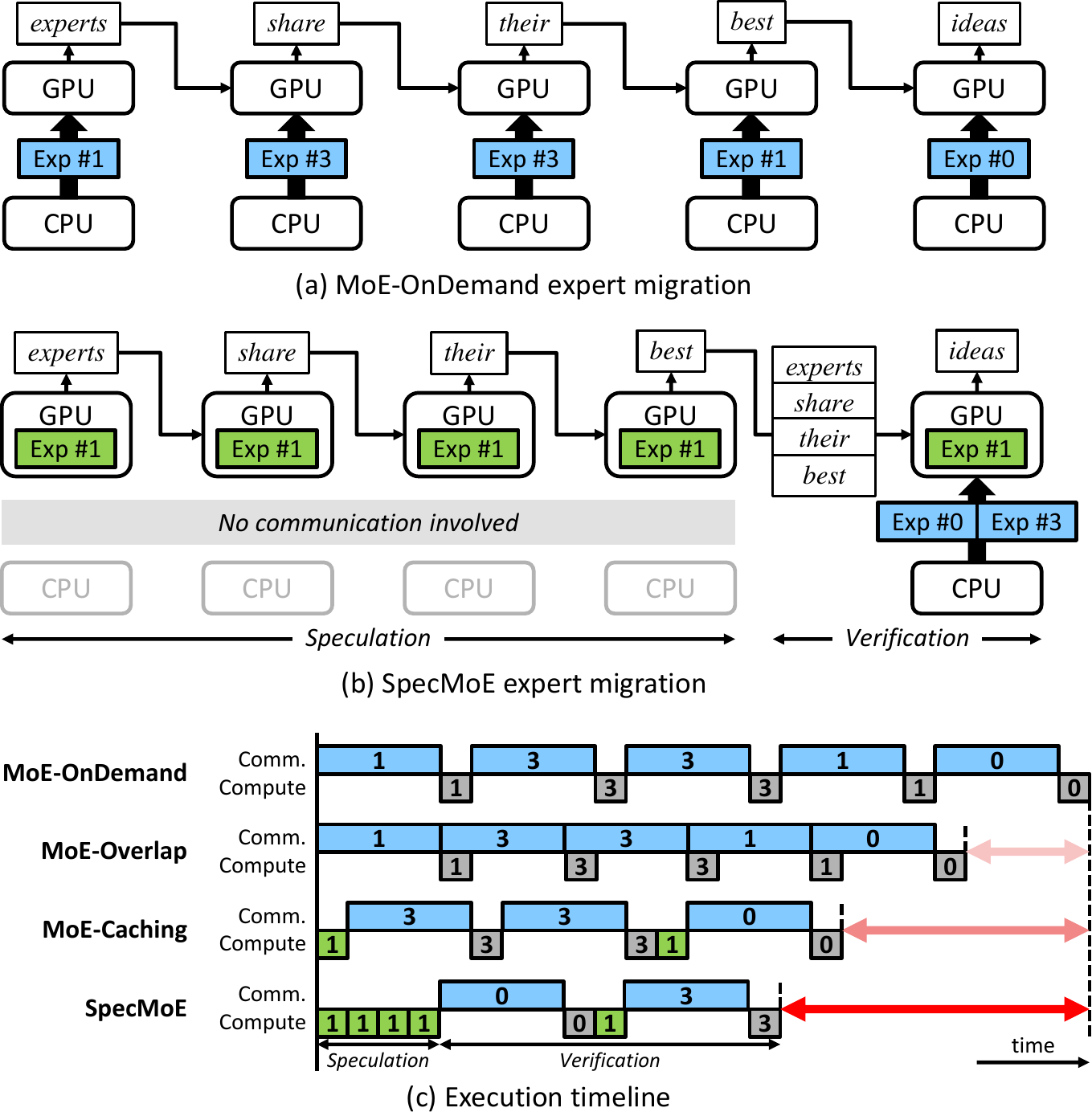}
    \caption{
    Comparison of data transfer size for expert migration between \ondemand and \proposed. \proposed reduces transfers by keeping frequently activated draft experts in GPU and minimizing redundant migrations. With $N=K=1$ (top-1 gating), \proposed retains \texttt{Exp \#1} in GPU, while \ondemand offloads all experts to CPU. In (a), \ondemand incurs CPU-to-GPU transfers every step. In (b), \proposed avoids migration in the speculation phase by generating all speculated tokens with \texttt{Exp \#1} in GPU. Migration occurs only in the verification phase when experts are accessed based on the gating function. Here, only \texttt{Exp \#0} and \texttt{Exp \#3} migrate once over 5 tokens, significantly reducing data transfer. (c) shows execution timelines, highlighting \proposed's efficiency.
    }
    \label{fig:benefit}
\end{figure}

\subsection{(System) CPU Offloading-based MoE Inference} \label{sect:system}

{\bf Speculative decoding for CPU offloaded systems.}
\proposed targets CPU offloading-based MoE inference systems, significantly reducing the volume of expert parameters migrated from CPU memory over PCIe.
We use the example in \fig{fig:benefit} to highlight \proposed's benefits by comparing it against \ondemand, but the key intuitions hold similarly for \overlap and \caching.
For \ondemand, as depicted in \fig{fig:benefit}(a), the activated experts must be migrated from CPU to GPU for each generated token.
Due to the auto-regressive nature of LLMs, tokens are generated sequentially and cannot be overlapped with each other.
Thus, even though \texttt{Exp \#1} and \texttt{Exp \#3} are used twice, their migrations cannot be coalesced and must happen separately, causing high communication overhead over PCIe.
However, in the \proposed system (\fig{fig:benefit}(b)), because the draft models do not require any expert migration, no communication is involved in the speculation phase.
During the verification phase, all expert migrations required by the target model are coalesced.
The verification of the four draft tokens occurs simultaneously and in parallel, amortizing the data transfer overhead of expert migration.
This approach enables \proposed to significantly reduce net data transfer compared to \ondemand by coalescing migration requests and minimizing the number of expert migrations.

It is worth emphasizing that batched LLM inference is becoming standard practice in LLM serving scenarios, due to the enormous cost associated with these large-scale generative AI model serving~\cite{ms_power_management, test_time_compute, splitwise}. This trend has led to the widespread adoption of high-performance batched LLM solutions like vLLM~\cite{vllm}.
These industry trends highlight the importance of \proposed, which delivers even greater performance benefits for batched inference scenarios compared to the three aforementioned CPU offloading-based MoE inference systems. Below we elaborate on \proposed's expected speed for batched speculative decoding-based inference.

{\bf Expected speedup with batched speculative decoding.}
As discussed in \sect{sect:background}, batched inference and speculative decoding pose their own challenges when applied to offload-based MoE systems, making them unfavorable choices in deployment.
However, we demonstrate how applying them \emph{simultaneously} holds great potential for speeding up offload-based MoE systems.
For a given batch size $B$ and draft token length of $\gamma$, let $\lambda$ denote the performance gap between (a) the latency incurred in the verification phase, $T_{\text{target}}^{<B\cdot\gamma>}$ (where the target model verifies $B\cdot\gamma$ tokens), and (b) the latency incurred in executing a single forward pass of the target model over the batch size $B$, $T_{\text{target}}^{<B>}$, so that $T_{\text{target}}^{<B\cdot\gamma>} = \lambda\cdot T_{\text{target}}^{<B>}$.
By applying this relationship to \equat{eq:speedup}, which quantifies the expected speedup with speculative decoding ($S$), we can re-write this equation to express the speedup $S$ in the batched inference context as follows:
\begin{align}
\label{eq:speedup_batch}
S = \frac{\tau(\gamma)\cdot T_{\text{target}}^{<B>}}{\gamma\cdot T_{\text{draft}}^{<B>} + T_{\text{target}}^{<B \cdot\gamma>}} = \frac{\tau(\gamma)}{\gamma\cdot c + \lambda}
\end{align}

Given this, consider a CPU offloading-based inference system where the expert migration latency generally dominates the end-to-end performance. In this context, $\lambda$ can be estimated by analyzing the data transfer size associated with expert migration, as it directly impacts $T_{\text{target}}^{<B \cdot \gamma>}$ and $T_{\text{target}}^{<B>}$ (as described in \sect{sect:challenge_specdec}).
Assuming the experimental setup shown in \fig{fig:latency_breakdown}(a), for a single-batch inference with $\gamma=8$, verifying $B \cdot \gamma$ tokens results in over six times more data transfers. This can be observed by comparing the data transfer sizes for batch sizes of 1 and 8 in \fig{fig:latency_breakdown}(a), leading to $\lambda \approx 6$.
However, this performance gap ($\lambda$) decreases significantly as the batch size $B$ increases. For instance, at $B=32$, verifying $B \cdot \gamma$ tokens requires only about twice as much data transfer, resulting in $\lambda \approx 2$. The key takeaway is that increasing the batch size $B$ significantly reduces $\lambda$, which in turn boosts the speedup $S$.
This makes our system design of batched speculative decoding an effective approach for CPU offloading-based MoE inference.

\begin{table}[t]
\centering
\vspace{0.6em}
\caption{MoE model configurations. For brevity, we only include information about the decoder blocks for NLLB-MoE. \emph{Skewness} is the portion of routed tokens covered by the top-25\% hottest experts. A higher skewness means the model has a more skewed expert activation distribution.}
\footnotesize
\resizebox{1.0\columnwidth}{!}{
\begin{tabular}{|c||c|c|c|c|c|}
\hline
\textbf{Model} & \textbf{Experts} & \textbf{\begin{tabular}[c]{@{}c@{}}Layers \\ (MoE / total)\end{tabular}} & \textbf{Routing} & \textbf{Skewness} & \textbf{\begin{tabular}[c]{@{}c@{}}Capacity \\ (GB)\end{tabular}} \\ \hline \hline
NLLB-MoE & 128 & 6 / 24 & Top-2 & 0.84 & \begin{tabular}[c]{@{}c@{}}54\end{tabular} \\ \hline
Mixtral-8x7B & 8 & 32 / 32 & Top-2 & 0.32 & \begin{tabular}[c]{@{}c@{}}94\end{tabular} \\ \hline
Llama-4-Scout & 16 & 48 / 48 & Top-1 & 0.59 & \begin{tabular}[c]{@{}c@{}}218\end{tabular} \\ \hline
\end{tabular}
}
\label{tab:model_dataset}
\end{table}

\section{Methodology} \label{sect:methodology}

{\bf Hardware and software.}
We evaluate \proposed on a CPU-GPU system with two Intel Xeon Platinum 8558 CPUs containing 1 TB of DDR5 CPU memory and a single NVIDIA H100 GPU with 96 GB of HBM3.
The CPU and GPU communicate over PCIe 5.0, providing 64 GB/s uni-directional communication bandwidth. Our software environment includes PyTorch 2.5.0 with CUDA 12.6, along with the Hugging Face Transformers 4.51.0 and Accelerate 1.6.0 libraries~\cite{hf_transformers, hf_accelerate}. The Hugging Face Transformers library's implementation of speculative decoding does not support batched inference and it also requires static draft model weights for execution. Because \proposed dynamically selects a subset of the target model's MoE block weights as its draft model, we develop a custom implementation of our self-assisted speculative decoding.

{\bf Model and dataset.}
We evaluate \proposed using three representative MoE models whose weights are publicly available (\tab{tab:model_dataset}). In \sect{sect:eval}, we focus our evaluation on NLLB-MoE~\cite{nllb} with the English–French translation task on the WMT-14~\cite{wmt} dataset. The effectiveness of \proposed on MoE models with less pronounced expert hotness is examined in \sect{sect:low_hotness}, where we evaluate Mixtral-8x7B~\cite{mixtral} and Llama-4-Scout~\cite{llama4} on the CNN-DailyMail~\cite{cnn_dailymail_nips, cnn_dailymail_acl} dataset for text completion.

{\bf Baseline system configurations.}
We evaluate three baseline systems: \ondemand, \overlap, and \caching.
All three baselines utilize the GPU memory to store the non-expert parameters while the sparse expert parameters are offloaded to the CPU memory as shown in \fig{fig:offloading_system}.
All inference-related computation is conducted using the GPU for all three baseline systems. However, \ondemand fetches activated experts from the CPU memory in an on-demand manner while \overlap overlaps expert migration with computations to hide the expert migration latency.
This is usually implemented by fetching the next layer's experts based on speculation or by modifying the model architecture as suggested in Pre-gated MoE~\cite{pregated_moe}. For a conservative evaluation, we establish an \emph{oracular} version of \overlap where the expert migration latency is completely overlapped with computation to maximize the expert migration latency hiding benefits of \overlap.
As for \caching, we keep the top 10\% most frequently accessed experts on GPU memory.

{\bf \proposed configuration.}
For \proposed's self-assisted speculative decoding, we utilize 4 draft experts per MoE block for constructing our draft model, configuring $\gamma$=10, where $\gamma$ represents the number of speculated draft tokens per speculation phase.
Using the greedy speculative decoding algorithm~\cite{blockwise}, \proposed's verification phase accepts only speculated draft tokens that precisely match the target model's greedy decoding results.

\begin{table}[t!]
\vspace{0.6em}
\caption{
Average number of generated tokens ($\tau(\gamma)$) for different expert selection policies (batch size: 256) and their effect on end-to-end inference performance (speedup normalized to {\bf Random}). $\gamma$=10 in our evaluations.
}
\centering
\renewcommand{\arraystretch}{1.0}
\resizebox{1.0\columnwidth}{!}{
\begin{tabular}{|c||c|c|c|c|}
\hline
{\bf Metric} & {\bf \randomexpert} & {\bf \hotglobal} & {\bf \hottemporal} \\
\hline
\hline
Tokens generated per step & 6.812 & 7.220 & 7.265 \\
\hline
Performance speedup & 1.000 & 1.084 & 1.143 \\
\hline
\end{tabular}
}
\label{tab:expert_seletion}
\end{table}

\section{Evaluation} \label{sect:eval}

The section focuses on the evaluation of NLLB-MoE, presenting the number of tokens generated per verification phase to show the robustness of our self-assisted speculative decoding algorithm (\sect{sect:tokens_per_step}).
We then evaluate the end-to-end inference latency and throughput (\sect{sect:performance}) to demonstrate \proposed's performance improvements, followed by a quantitative analysis of the effect of \proposed on CPU-to-GPU data transfers (\sect{sect:efficiency_and_scalability}), which explains the primary reasons behind \proposed's performance gains.

\subsection{Generated Tokens Per Verification Phase} \label{sect:tokens_per_step}

The key to \proposed's success is maximizing the number of speculated tokens actually accepted during verification.
If too few of the speculated tokens are accepted, the time spent speculating those tokens is essentially wasted, resulting in slower generation speeds compared to other baseline systems.
Furthermore, during the verification phase, \proposed retrieves the activated experts for all speculated tokens, regardless of whether those tokens are eventually accepted or not.
This may lead to unnecessary migration of experts even for incorrectly speculated tokens, causing a miss penalty that aggravates the CPU-to-GPU communication bottleneck.

In \tab{tab:expert_seletion}, we explore three different expert selection policies for the draft model to determine their effect on the speculated token acceptance rate.
\emph{Tokens generated per step} refers to the number of tokens generated within a single speculative decoding step.
That is, it includes both the number of speculated tokens and the resampled token during the verification phase (\fig{fig:spec_timeline}).
Under the \textbf{\randomexpert} policy, the draft experts are chosen randomly from each MoE block.
\textbf{\hotglobal} identifies the \emph{hot experts} that are accessed most frequently (e.g., \fig{fig:hotness}) within each MoE block, and statically chooses them as the draft experts.
\textbf{\hottemporal} is the policy in use by our \proposed algorithm, which dynamically captures and replaces temporally hot experts during each verification phase.
According to our evaluation, \randomexpert generates 6.812 tokens on average per verification phase.
This result shows the algorithmic robustness of our self-assisted draft model design approach, even under a random selection policy.
With access to the expert activation hotness information, \hotglobal generates 7.220 tokens.
Finally, by taking the temporal shift in expert activation hotness into account, \hottemporal generates up to 7.265 tokens per speculative decoding step, demonstrating the effectiveness of \proposed in improving token generation performance, providing an additional 6\% end-to-end speedup over \hotglobal.
By enabling the draft model to generate speculated tokens closely aligned with the target model, \proposed with \hottemporal enhances the overall performance.

\subsection{Performance} \label{sect:performance}

\begin{figure}
    \centering
    \includegraphics[width=0.9\columnwidth]{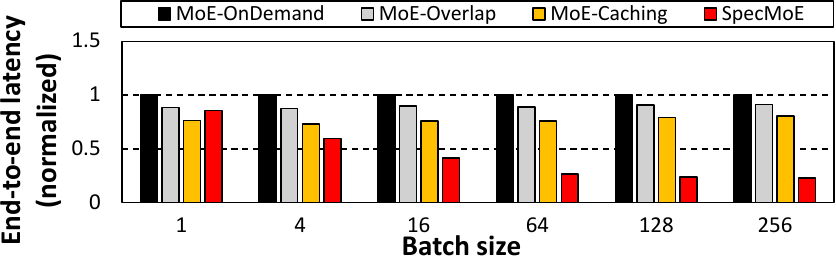}
    \caption{
    Normalized end-to-end inference latency (NLLB-MoE).
    }
    \label{fig:eval_latency_nllb}
\end{figure}

{\bf End-to-end inference latency.}
\fig{fig:eval_latency_nllb} compares \proposed's end-to-end latency vs. the three baselines over a batch size ranging from 1 to 256. While today's LLM inference systems are optimized for large batches (underscoring the importance of system-level solutions like vLLM~\cite{vllm} and others~\cite{orca, s_lora, dlora, punica, splitwise}), we nonetheless include extremely low batch inference scenarios like batch size of 1 for the completeness of our study.
\proposed's latency reduction benefit becomes proportionally higher as batch size increases, incurring only 23\% of the \ondemand's latency at the batch size of 256. These results highlight \proposed's ability to minimize CPU-to-GPU expert migration traffic, which is the primary bottleneck of CPU offloading-based MoE inference over large batch sizes.

\begin{figure}[t]
    \centering
    \includegraphics[width=0.9\columnwidth]{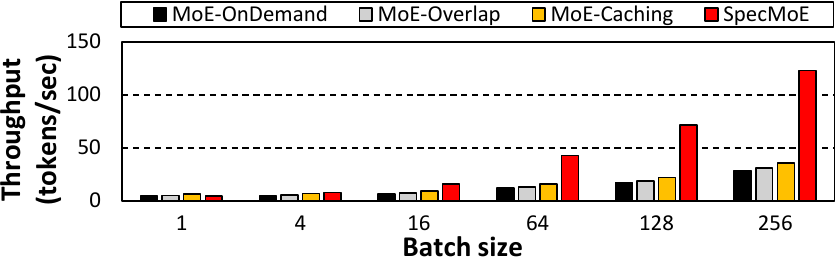}
    \caption{
    End-to-end inference throughput (tokens per second).
    }
    \label{fig:eval_throughput_nllb}
\end{figure}

It is worth noting that \caching show slightly shorter end-to-end latency than \proposed under the batch size 1 inference scenario.
This is because the model activates only 2 out of the 128 total experts per MoE block under batch 1 inference, and \caching likely has these experts already loaded in GPU memory, as it caches 10\% of the total experts on the GPU.
Nevertheless, addressing \proposed's inefficiency under low-batch inference scenarios in real-world LLM inference systems is straightforward. This is because disabling \proposed's speculative decoding feature under low batch sizes can be easily implemented at the software level without any modifications to the MoE model or causing additional hardware resource usage.
This ensures \proposed's seamless adaptability for real LLM serving scenarios while preserving our proposal's significant advantage under large-batch inference settings.

{\bf Token generation throughput.}
We also show the changes in inference throughput (measured using the number of tokens generated per second, \fig{fig:eval_throughput_nllb}) as batch sizes change.
All evaluated systems generally achieve an increase in inference throughput as batch size is increased.
However, the rate in which throughput is increased is the highest and most pronounced with \proposed, achieving up to 123.0 tokens/sec and outperforming all other design points that exhibit only limited throughput improvements.
This is because the baseline systems can only partially mitigate the expert migration latency, showing limited effectiveness in large batch inference. Under large batches where a larger number of experts are activated and fetched from CPU to GPU, the three baseline systems exhibit a near-linear increase in latency, as shown in \fig{fig:latency_breakdown}, limiting its throughput increase with large batches.

Unlike these baseline systems, \proposed demonstrates a proportional increase in throughput with larger batch sizes, achieving up to a 4.30$\times$ speedup over \ondemand. This improvement is due to \proposed delaying all expert migration requests for draft tokens until the verification phase. By doing so, experts that are activated multiple times during the speculation and verification phases are migrated exactly once, eliminating redundant expert migrations (\fig{fig:benefit}). In summary, the significant throughput improvements achieved by \proposed result from its efficient handling of data transfers and its ability to generate multiple tokens per single target model execution. The details of this reduction in data transfer will be examined further in the following section.

\begin{figure}[t]
    \centering
    \includegraphics[width=0.9\columnwidth]{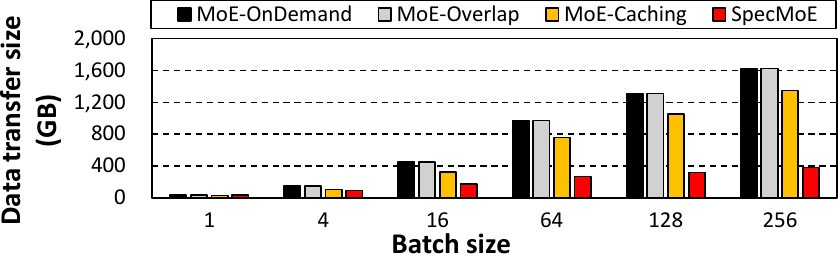}
    \caption{
    CPU-to-GPU data transfer size in end-to-end inference.
    }
    \label{fig:eval_data_transfer_nllb}
\end{figure}

\subsection{Efficiency and Scalability} \label{sect:efficiency_and_scalability}

\fig{fig:eval_data_transfer_nllb} compares the size of expert parameters migrated during the CPU-to-GPU communications.
\ondemand fetches required experts from the CPU to the GPU at every decoding step, resulting in substantial data transfers.
Similarly, while \overlap tries to hide the migration latency by overlapping it with computation time, it still needs to transfer the same amount of expert parameters over PCIe, resulting in no improvement over \ondemand in terms of data transfer size.
\caching can reduce expert migration traffic compared to the other two baselines by caching the frequently accessed experts in the GPU memory.
However, \caching's reduction in data transfer size for the non-cached experts is almost non-existent, resulting in \caching to achieve a moderate 30\% reduction in overall data transfer size.
\proposed, on the other hand, directly tackles the problem of excessive data transfers, minimizing the CPU-to-GPU communication traffic by up to 76.73\% compared to \ondemand and \overlap, and by 71.89\% compared to \caching.
Interestingly, \proposed achieves such high reduction in data transfer size while using much less GPU memory than \caching, which caches 13 (10\% of 128) hot experts while \proposed caches only 4 experts to construct its draft model.

The advantages of \proposed become clearer as the batch size increases.
The three baseline systems experience sharp increases in data transfer size with larger batches; for instance, when increasing the batch size from 1 to 256, the data transfer size grows 45.28$\times$ for MoE-OnDemand/MoE-Overlap and 50.97$\times$ for \caching.
\proposed achieves a much smaller increase of only 10.96$\times$ during the same batch size increase interval.
\proposed's efficiency in minimizing data transfers stems from our self-assisted speculative decoding algorithm, where multiple tokens are validated altogether during the verification phase.
As the batch size grows, the number of tokens processed simultaneously in the verification phase grows significantly.
Consequently, the likelihood of these tokens being routed to identical experts also increases, reducing redundant expert migrations and further optimizing data transfer size.
In summary, unlike other baselines that migrate the same experts repeatedly, \proposed optimizes the data transfer between CPU and GPU, demonstrating superior scalability as the batch size grows.

\section{Discussion} \label{sect:discussion}

\begin{figure}[t]
    \centering
    \includegraphics[width=0.96\columnwidth]{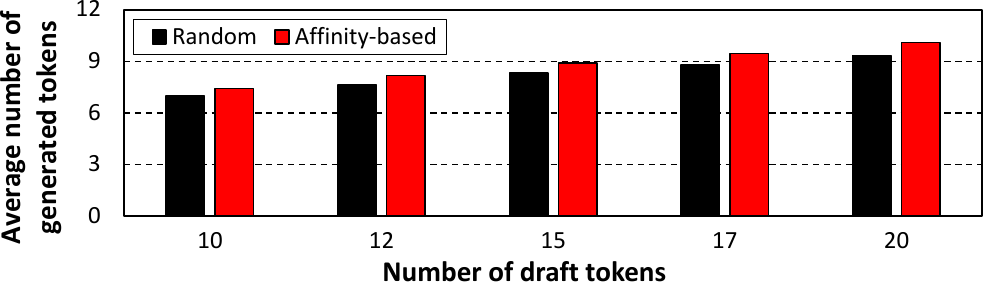}
    \caption{Average number of tokens generated per speculative decoding step as the number of draft tokens varies (batch size: 128).}
    \label{fig:affine_ablation}
\end{figure}

\subsection{Effect of Affinity-based Expert Selection}

As explained in \sect{sect:algo}, \proposed refines the routing results from the gating function using the precomputed affinity table during the speculation phase.
To evaluate the effectiveness of our affinity-based expert selection mechanism, we perform an ablation study on the expert selection policy. \fig{fig:affine_ablation} illustrates the average number of tokens generated per speculative decoding step ($\tau(\gamma)$) as the number of draft tokens ($\gamma$) varies from 10 to 20. We compare two approaches: randomly selecting experts (Random) and using our affinity-based expert selection (Affinity-based).
As shown in the figure, the affinity-based approach consistently outperforms the random selection, generating 7.23\% more tokens per step on average.
These results provide compelling evidence that our affinity-based expert selection method effectively approximates the original computation by selecting experts that are most similar to those to which tokens should be routed, thereby leading to high-quality draft generation.

\subsection{Effect of Draft Expert Number on \proposed} \label{sect:different_n_experts}

\begin{table}[t!]
\vspace{0.6em}
\caption{
Average number of tokens generated per step ($\tau(\gamma)$) and achieved throughput for $\gamma=10$ when the number of draft experts ($N$) is changed in NLLB-MoE (batch size: 256).
}
\centering
\renewcommand{\arraystretch}{1.0}
\small
\resizebox{1.0\columnwidth}{!}{
\begin{tabular}{|c||>{\centering\arraybackslash}m{1cm}|>{\centering\arraybackslash}m{1cm}|>{\centering\arraybackslash}m{1cm}|>{\centering\arraybackslash}m{1cm}|}
\hline
\multirow{2}{*}{\textbf{Metric}} & \multicolumn{4}{c|}{\textbf{Number of experts in draft model ($N$)}} \\ \cline{2-5}
                        & 2           & 4           & 8           & 16          \\ \hline \hline
Tokens generated  per step         & 6.929       & 7.265       & 7.620       & 7.819       \\ \hline
Throughput (tokens/sec)         & 116.401       & 122.974       & 121.309       & 117.157       \\ \hline
\end{tabular}
}
\label{tab:draft_experts}
\end{table}

This section evaluates how the number of experts in the draft model affects \proposed performance.
\tab{tab:draft_experts} presents the number of tokens generated per speculative decoding step as the number of draft experts ($N$) varies from 2 to 16 (our evaluation in \sect{sect:eval} used 4 draft experts per MoE block for NLLB-MoE). As shown, the number of generated tokens increases with a larger number of draft experts. However, the end-to-end inference throughput does not scale proportionally with the number of tokens generated per step and even declines in certain cases. When more experts are included in the draft model (i.e., as $N$ increases), the latency to execute these experts also rises. This results in lower compute utilization because the number of tokens processed per step remains the same. Therefore, a larger $N$ does not always lead to better end-to-end performance. To maximize efficiency, \proposed should identify the optimal value of $N$, balancing the acceptance of more speculated tokens against the cost of sequential computations in the draft MoE blocks. We therefore chose $N$=4 as NLLB-MoE's draft model in \sect{sect:eval} to balance draft model quality and end-to-end performance.

\subsection{\proposed Effectiveness on MoE Models with Low Expert Activation Hotness} \label{sect:low_hotness}

For the completeness of our work, we evaluate an MoE model that exhibits a relatively uniform expert activation distribution: Mixtral-8x7B~\cite{mixtral} and Llama-4-Scout~\cite{llama4}. To measure how uniform (or how skewed) an MoE model is, we define \emph{skewness} of an MoE model as the portion of tokens that are routed to the \emph{top 25\%} frequently activated experts, averaged throughout all layers. Skewness is a value between 0.25 and 1 where the closer it is to 0.25, the more uniformly activated its experts are. As reported in \tab{tab:model_dataset}, NLLB-MoE is highly skewed and is biased to activate certain hot experts.
In contrast, Mixtral-8x7B and Llama-4-Scout, with a skewness of 0.32 and 0.59, respectively, are models whose experts are relatively uniformly activated compared to NLLB-MoE.
Because expert activation hotness is not strongly observed, it becomes more challenging for \proposed's draft model to include the optimal set of draft experts. This results in a lower draft token acceptance rate compared to MoE models with skewed expert hotness.
To address this, we set the number of speculated draft tokens ($\gamma$) to 5 for these models. This lower value of $\gamma$, compared to NLLB-MoE's $\gamma$=10, compensates for the reduced number of generated tokens per step. A higher $\gamma$ value in such case would unnecessarily increase speculation overhead without significant improvements in the number of tokens generated per step ($\tau(\gamma)$).

\begin{figure}[t]
    \centering
    \includegraphics[width=0.9\columnwidth]{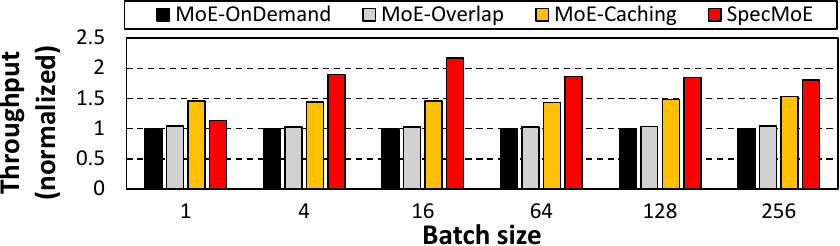}
    \caption{End-to-end inference throughput of Mixtral-8x7B.}
    \label{fig:throughput_mixtral}
\end{figure}

\begin{figure}[t]
    \centering
    \includegraphics[width=0.9\columnwidth]{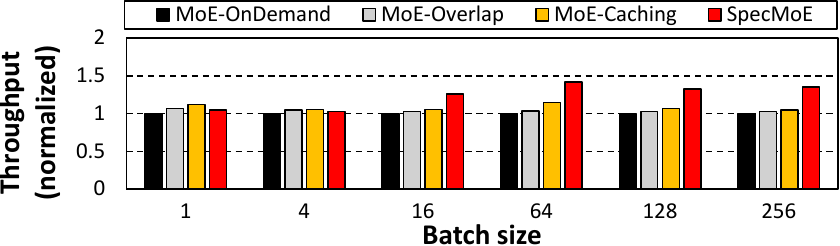}
    \caption{End-to-end inference throughput of LLama-4-Scout.}
    \label{fig:throughput_llama}
\end{figure}

With this design decision, we configure the verification phase of \proposed to use the sampling-based methodology introduced in \cite{specdec}. We perform the text completion task on the CNN-DailyMail dataset with a temperature of 1.
As a result, \proposed presents slightly lower performance than \caching for small batch sizes (less than or equal to 4) similar to the results observed in \fig{fig:eval_throughput_nllb}.
Nevertheless, for batch sizes larger than 4, \proposed consistently outperforms the three baseline CPU offload–based systems.
As shown in \fig{fig:throughput_mixtral} and \fig{fig:throughput_llama}, \proposed achieves up to 2.17$\times$ and 1.42$\times$ improvements in throughput over \ondemand for Mixtral-8x7B and Llama-4-Scout, respectively. By contrast, \overlap and \caching deliver maximum improvements of only 1.53$\times$ and 1.15$\times$ for Mixtral-8x7B and Llama-4-Scout, respectively.
These results highlight that \proposed delivers noticeable performance gains even for MoE models that do not exhibit a skewed distribution of expert activation hotness, underscoring its robust design.

Importantly, \proposed achieves this speedup without additional resources or modifications to the target model architecture.
Recent best practices for draft-model development select models roughly 1\% the size of the target~\cite{specdec, specinfer, specexec, sequoia, staged_spec, sambanova-specdec}.
Applying this ratio to Mixtral-8$\times$7B and Llama-4-Scout still requires draft models with several billion parameters.
Generating draft tokens with such draft models consumes significant GPU memory for both the draft model weights and associated KV caches, which is especially critical in resource-constrained, offloading-based inference systems.
Furthermore, selecting an appropriate draft model is a challenging task. For example, Llama-4-Scout is the smallest model of the Llama-4 family (i.e., there is no smaller model that uses the same tokenizer). Thus, applying speculative decoding to this model would require creating a new, smaller model or adopting drafter-free approaches, both of which demand modifications to the target model architecture and additional efforts for training.
In contrast, \proposed employs a self-assisted speculative decoding approach that requires no extra GPU memory for draft model parameters or KV caches.

\subsection{\proposed with SSD-Offloading}

\begin{figure}[t]
    \centering
    \includegraphics[width=0.96\columnwidth]{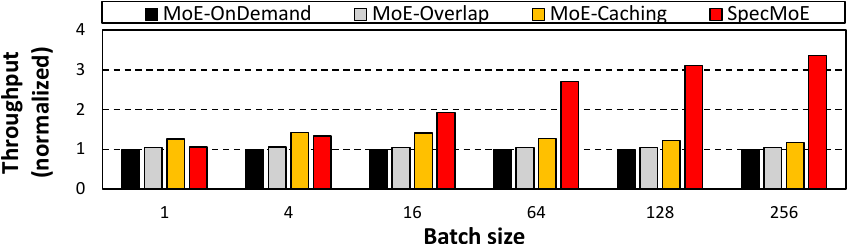}
    \caption{Normalized end-to-end inference throughput of NLLB-MoE when expert parameters are offloaded to the SSD.}
    \label{fig:throughput_ssd}
\end{figure}

Prior work~\cite{se_moe, pregated_moe} explored the efficacy of offloading MoE experts to SSDs as a means to deploy even larger LLMs.
SSD-offloading introduces higher data transfer latency due to the limited I/O bandwidth of SSDs, making expert migration latency dominate the end-to-end execution. As shown in \fig{fig:throughput_ssd}, \proposed achieves on average a 2.25$\times$ throughput improvement over \ondemand, while \overlap and \caching demonstrate only 1.05$\times$ and 1.29$\times$ average speedups, respectively.
The subpar performance of these two baselines stems from the limited I/O bandwidth, which exacerbates expert migration latency and reduces the effectiveness of latency overlapping and expert caching.
Because \proposed fundamentally minimizes the data transfer size by eliminating redundant expert migrations, it is able to consistently provide speed boosts compared even when the experts are offloaded to SSDs.
These results highlight \proposed's utility in high-latency, bandwidth-constrained environments such as SSDs.

\section{Related Work}

{\bf Offload-based systems.}
ZeRO-offload~\cite{zero_offload} and ZeRO-infinity~\cite{zero_infinity} offload model parameters to slower-tier memory, enabling large AI model training. FlexGen~\cite{flexgen} introduces a scheduling algorithm to maximize throughput in GPU-constrained, offloading-based environments. However, these approaches do not account for sparse activation and are therefore not applicable to MoE models. Our implementations of \ondemand and \overlap from \sect{sect:moe} are based on Hugging Face Accelerate~\cite{hf_accelerate} and Pre-gated MoE~\cite{pregated_moe}, respectively. SE-MoE~\cite{se_moe} and Huang et al.~\cite{meta_moe} each propose expert-offloading schemes for MoE using methods similar to those described for \overlap and \caching, respectively. However, these methods do not address the data transfer bottleneck during batched inference, a challenge addressed in our \proposed.
PowerInfer~\cite{powerinfer} characterizes column-/row-wise activation sparsity in weight parameters and leverages this insight to offload cold parts of the weight parameters to the CPU while retaining hot parts on the GPU, reducing GPU memory usage and data transfers.
However, this approach is less effective to models that do not use ReLU functions, which is common in state-of-the-art LLMs.

{\bf Efficient MoE serving.}
To address the inefficiency inherent in the sparse activation patterns of MoE, DeepSpeed-Inference~\cite{deepspeed_inference} and SE-MoE~\cite{se_moe} propose new model architectures based on knowledge distillation. However, these approaches incur additional training costs and result in model accuracy degradation. Duplex~\cite{duplex} introduces a hardware accelerator for batched MoE inference but its applicability is limited to a specific type of attention mechanism, grouped query attention~\cite{gqa}. \proposed incurs no additional training costs, maintains accuracy, and is applicable to any MoE model.

{\bf Speculative decoding.}
Blockwise~\cite{blockwise}, Leviathan et al.~\cite{specdec}, and SpS~\cite{sps} introduced the concept of speculative decoding applied in our work.
Several prior work proposes to store speculated tokens in tree-like structures, with \cite{staged_spec} applying a new speculative decoding scheme to the draft models, SpecInfer~\cite{specinfer} introducing hardware-friendly tree attention mechanism, and Sequoia~\cite{sequoia} optimizing and deepening the tree-like structure.
SpecExec~\cite{specexec} introduces tree-based speculation with a verification algorithm that caches and reuses the target model's logits, demonstrating this approach's efficacy on consumer GPUs. EdgeLLM~\cite{edgellm} proposes a compute-efficient token-tree generation mechanism and an inference pipeline that continues speculation during verification, enabling speculative decoding on edge devices.
Overall, these works require the help of a separately trained draft model, requiring additional compute resources, and are limited to applications to dense models.
\proposed proposes a training-free draft model for MoEs.

In an attempt to reduce the training cost of the draft model, EAGLE~\cite{eagle} reuses the target model's embedding layer and LM head. By focusing on the output logit distribution rather than the output token, it is able to quickly retrain the draft model to achieve a moderate draft model quality.
Nonetheless, Eagle still requires retraining to maintain its output generation quality, while the draft model proposed in \proposed is completely training-free.

LayerSkip~\cite{layerskip}, Speed~\cite{speed}, and Draft\&Verify~\cite{draft_verify} reuse parts of the target model to create the draft model and use early exit mechanisms to accelerate generation speed and reduce draft model training costs. These techniques, however, are generally designed for dense LLMs and cannot be directly applied to MoE models. Additionally, LayerSkip requires retraining of the target model.
Medusa~\cite{medusa} eliminates the need for a separate draft model by adding multiple LM heads at the end of target dense LLMs, enabling the generation of multiple output tokens in a single forward pass. However, this approach still incurs additional compute overhead to align the LM heads with the model.
In contrast, \proposed requires no retraining and is compatible with all MoE architectures.

\section{Conclusion}

\proposed proposes self-assisted speculative decoding to address the challenges of CPU-to-GPU communication overhead in CPU offloading-based MoE inference.
By composing the draft model from parts of the target model and dynamically reconstructing it at runtime, the speculation phase of \proposed generates quality tokens while residing entirely in the GPU.
The verification phase of \proposed alleviates the data transfer overhead by coalescing any duplicate expert migrations, effectively removing all redundant transfers over the PCIe channel.
On top of that, \proposed avoids the need for any separate draft model training, saving on training compute costs.
Overall, \proposed is an efficient hardware-aware, software-only solution that incurs no additional training cost that can be generally applied to any MoE serving systems.

\bibliographystyle{IEEEtranS}
\bibliography{refs}

\end{document}